\documentclass[10pt,twocolumn,letterpaper]{article}

\usepackage{style/cvpr}
\usepackage{times}
\usepackage{epsfig}
\usepackage{graphicx}
\usepackage{amsmath}
\usepackage{amssymb}
\usepackage{booktabs}
\usepackage{xcolor}
\usepackage{multirow}
\usepackage{authblk}
\usepackage{capt-of}
\newcommand\ignorethis[1]{}
\newcommand\KG[1]{}
\newcommand\TF[1]{}
\newcommand\AS[1]{}

\usepackage[pagebackref=true,breaklinks=true,letterpaper=true,colorlinks,bookmarks=false]{hyperref}

\cvprfinalcopy %

\def\cvprPaperID{7472} %

\ifcvprfinal\fi
\begin{document}

\title{Local Deep Implicit Functions for 3D Shape\vspace*{-0.1in}}

\makeatletter
\renewcommand\Authfont{\fontsize{11.5}{14.4}\selectfont}
\renewcommand\AB@affilsepx{\qquad \protect\Affilfont}
\makeatother
\author[1,2]{Kyle Genova}
\author[2]{Forrester Cole}
\author[2]{Avneesh Sud}
\author[2]{Aaron Sarna}
\author[1, 2]{Thomas Funkhouser}
\affil[1]{Princeton University}
\affil[2]{Google Research}
\renewcommand*{\Authsep}{ \ }%
\renewcommand*{\Authands}{ \ }%

\twocolumn[{%
\renewcommand\twocolumn[1][]{#1}%
\maketitle
\includegraphics[width=\textwidth]{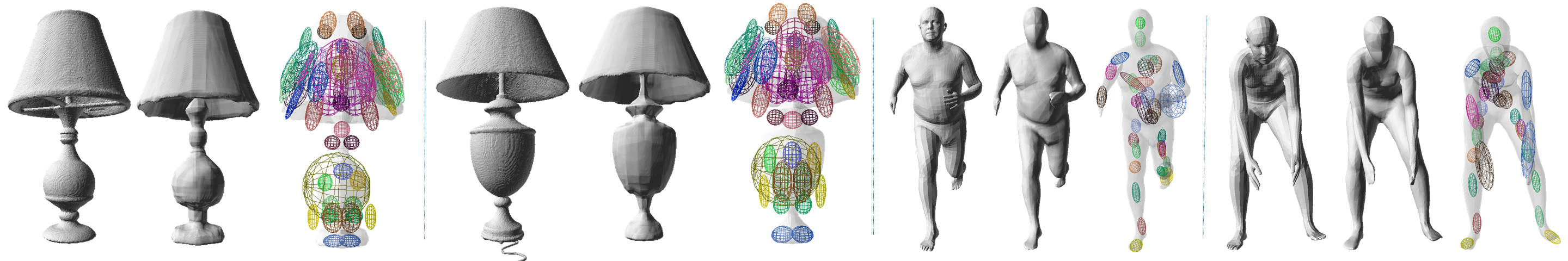}
\captionof{figure}{This paper introduces Local Deep Implicit Functions, a 3D shape representation that decomposes an input shape (mesh on left in every triplet) into a structured set of shape elements (colored ellipses on right) whose contributions to an implicit surface reconstruction (middle) are represented by latent vectors decoded by a deep network. Project video and website at \href{https://ldif.cs.princeton.edu}{ldif.cs.princeton.edu}. \vspace*{1cm}}
\label{fig:teaser}
}]

\maketitle

\setlength{\abovedisplayskip}{0.5\abovedisplayskip}
\setlength{\belowdisplayskip}{0.5\belowdisplayskip}
\setlength{\abovecaptionskip}{0.5\abovecaptionskip}
\setlength{\belowcaptionskip}{0.5\belowcaptionskip}

\begin{abstract}
The goal of this project is to learn a 3D shape representation that enables accurate surface reconstruction, compact storage, efficient computation, consistency for similar shapes, generalization across diverse shape categories, and inference from depth camera observations.  Towards this end, we introduce Local Deep Implicit Functions (LDIF), a 3D shape representation that decomposes space into a structured set of learned implicit functions.  We provide networks that infer the space decomposition and local deep implicit functions from a 3D mesh or posed depth image.  During experiments, we find that it provides 10.3 points higher surface reconstruction accuracy (F-Score) than the state-of-the-art (OccNet), while requiring fewer than 1\% of the network parameters.  Experiments on posed depth image completion and generalization to unseen classes show 15.8 and 17.8 point improvements over the state-of-the-art, while producing a structured 3D representation for each input with consistency across diverse shape collections.

\end{abstract}
\section{Introduction}

Representing 3D shape is a fundamental problem with many applications, including surface reconstruction, analysis, compression, matching, interpolation, manipulation, and visualization.   For most vision applications, a 3D representation should support: (a) reconstruction with accurate surface details, (b) scalability to complex shapes, (c) support for arbitary topologies, (d) generalizability to unseen shape classes, (e) independence from any particular application domain, (f) encoding of shape priors, (g) compact storage, and (h) computational efficiency.   

No current representation has all of these desirable properties.
Traditional explicit 3D representations (voxels, meshes, point clouds, etc.) provide properties (a-e) above.  They can represent arbitrary shapes and any desired detail, but they don't encode shape priors helpful for efficient storage, 3D completion, and reconstruction tasks. In contrast, learned representations (latent vectors and deep network decoders) excel at representing shapes compactly with low-dimensional latent vectors and encoding shape priors in network weights, but they struggle to reconstruct details for complex shapes or generalize to novel shape classes.

Most recently, deep implicit functions (DIF) have been shown to be highly effective for reconstruction of individual objects~\cite{mescheder2019occ, chen2018learning, park2019deepsdf, xu2019disn}. They represent an input observation as a latent vector $\mathbf{z}$ and train a neural network to estimate the inside/outside or signed-distance function $f(\mathbf{x}, \mathbf{z})$ given a query location $\mathbf{x}$ in 3D space. This approach achieves state of the art results for several 3D shape reconstruction tasks. However, they use a single, fixed-length latent feature vector to represent the entirety of all shapes and they evaluate a complex deep network to evaluate the implicit function for every position $\mathbf{x}$.   As a result, they support limited shape complexity, generality, and computational efficiency.

Meanwhile, new methods are emerging for learning to infer structured decomposition of shapes~\cite{tulsiani2017learning, genova2019learning}.  For example,~\cite{genova2019learning} recently proposed a network to encode shapes into Structured Implicit Functions (SIF), which represents an implicit function as a mixture of local Gaussian functions.  They showed that simple networks can be trained to decompose diverse collections of shapes consistently into SIFs, where the local shape elements inferred for one shape (e.g., the leg of a chair) correspond to semantically similar elements for others (e.g., the leg of a table).   However, they did not use these structured decompositions for accurate shape reconstruction due to the limited shape expressivity of their local implicit functions (Gaussians).

The key idea of this paper is to develop a pipeline that can learn to infer {\em Local Deep Implicit Functions}, (``LDIF'', Figure 1).  An LDIF is a set of local DIFs that are arranged and blended according to a SIF template. The representation is similar to SIF in that it decomposes a shape into a set of overlapping local regions represented by Gaussians. However, it also associates a latent vector with each local region that can be decoded with a DIF to produce finer geometric detail.   
Alternately, LDIF is similar to a DIF in that it encodes a shape as a latent vector that can be evaluated with a neural network to estimate the inside/outside function $f(\mathbf{x}, \mathbf{z})$ for any location $\mathbf{x}$. However, the LDIF latent vector is decomposed into parts associated with local regions of space (SIF Gaussians), which makes it more scalable, generalizable, and computationally efficient.

In this paper, we not only propose the LDIF representation, but we also provide a common system design that works effectively for 3D autoencoding, depth image completion, and partial surface completion.  First, we propose to use DIF to predict local functions that are \emph{residuals} with respect to the Gaussian functions predicted by SIF -- this choice simplifies the task of the DIF, as it must predict only fine details rather than the overall shape within each shape element.   Second, we propose to use the SIF decomposition of space to focus the DIF encoder on local regions by gathering input 3D points within each predicted shape element and encoding them with PointNet \cite{qi2016pointnet}.  Finally, we investigate several significant improvements to SIF (rotational degrees of freedom, symmetry constraints, etc.) and simplifications to DIF (fewer layers, smaller latent codes, etc.) to improve the LDIF representation.  Results of ablation studies show that each of these design choices provides significant performance improvements over alternatives.   In all, LDIF achieves 10-15 points better F-Score performance on shape reconstruction benchmarks than the state-of-the-art \cite{mescheder2019occ}, with fewer than 1\% of the network parameters.

\section{Related Work}

\noindent{\bf Traditional Shape Representations:} There are many existing approaches for representing shape. In computer graphics, some of the foundational representations are meshes~\cite{baumgart1975polyhedron}, point clouds~\cite{foley1996computer}, voxel grids~\cite{foley1996computer}, and implicit surfaces~\cite{ricci1973constructive, blinn1982generalization, bloomenthal1991convolution, muraki1991volumetric, ohtake2003multi, wyvill1986soft}. These representations are popular for their simplicity and ability to operate efficiently with specialized hardware. However, they lack two important properties: they do not leverage a shape prior, and they can be inefficient in their expressiveness. Thus, traditional surface reconstruction pipelines based on them, such as Poisson Surface Reconstruction~\cite{kazhdan2006poisson}, require a substantial amount of memory and computation and are not good for completing unobserved regions.

\vspace*{1mm}\noindent{\bf Learned Shape Representations:} To leverage shape priors, shape reconstruction methods began representing shape as a learned feature vector, with a trained decoder to a mesh~\cite{smith2019geometrics, gkioxari2019mesh, wang2018pixel2mesh, groueix2018papier, kanazawa2018learning}, point cloud~\cite{fan2017point, lin2018learning, yang2019pointflow}, voxel grid~\cite{choy20163d, wu20153d, brock2016generative, wu2016learning}, or octree \cite{tatarchenko2017octree, riegler2017octnetfusion, riegler2017octnet}. Most recently, representing shape as a vector with an implicit surface function decoder has become popular, with methods such as OccNet~\cite{mescheder2019occ}, ImNet~\cite{chen2018learning}, DeepSDF~\cite{park2019deepsdf}, and DISN~\cite{xu2019disn}. These methods have substantially improved the state of the art in shape reconstruction and completion. However, they do not scale or generalize very well because the fundamental representation is a single fixed-length feature vector representing a shape globally.   

\vspace*{1mm}\noindent{\bf Structured Shape Representations:}  To improve scalability and efficiency, researchers have introduced structured representations that encode the repeated and hierarchical nature of shapes. Traditional structured representations include scene graphs~\cite{foley1996computer}, CSG trees~\cite{foley1996computer}, and partition of unity implicits~\cite{ohtake2003multi}, all of which represent complex shapes as the composition of simpler ones.  Learned representations include SDM-Net~\cite{gao2019sdm}, GRASS~\cite{li2017grass}, CSGNet~\cite{sharma2018csgnet}, Volumetric Primitives~\cite{tulsiani2017learning}, Superquadrics~\cite{Paschalidou2019CVPR}, and SIF~\cite{genova2019learning}.  These methods can decompose shapes into simpler ones, usually with high consistency across shapes in a collection. However, they have been used primarily for shape analysis (e.g. part decomposition, part-aware correspondence), not for accurate surface reconstruction or completion.  Concurrent work~\cite{chibane2020implicit, jiang2020local} adds a voxel grid structure to deep implicit functions. This helps preserve local detail but does not take advantage of consistent shape decomposition.

\begin{figure*}
    \centering
    \includegraphics[width=\textwidth]{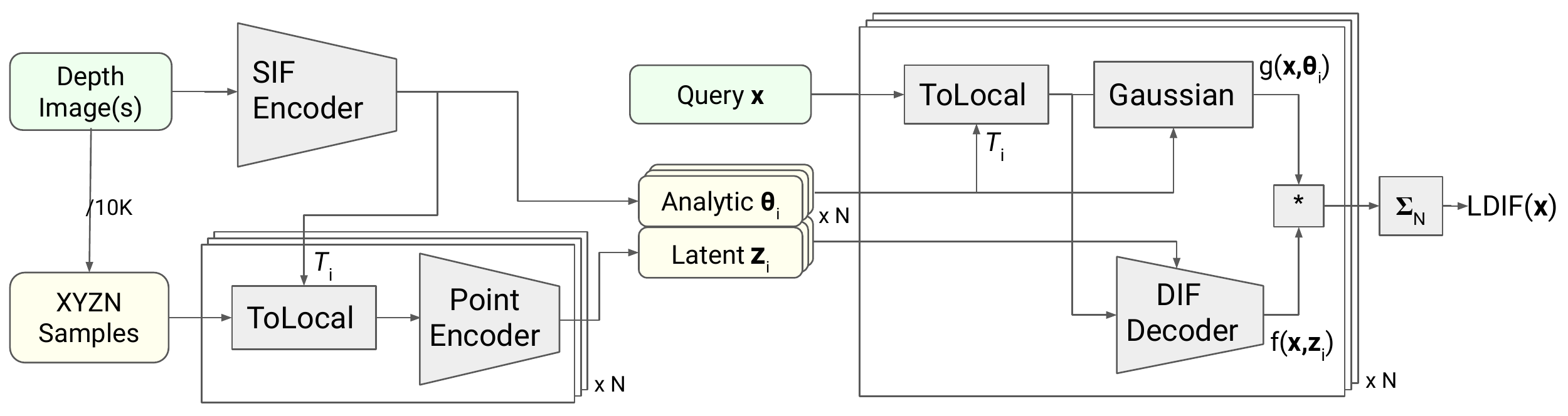}
    \caption{\textbf{Network architecture.} Our system takes in one or more posed depth images and outputs an LDIF function that can be used to classify inside/outside for any query point $\mathbf{x}$.   It starts with a SIF encoder to extract a set of overlapping shape elements, each defined by a local Gaussian region of support parameterized by $\mathbf{\theta_i}$.  It then extracts sample points/normals from the depth images and passes them through a PointNet encoder for each shape element to produce a latent vector $\mathbf{z_i}$.  A local decoder network is used to decode each $\mathbf{z_i}$ to produce an implicit function $f_i(\mathbf{x}, \mathbf{z_i})$, which is combined with the local Gaussian function $g(\mathbf{x}, \mathbf{\theta_i})$
    and summed with other shape elements to produce the output function LDIF$(\mathbf{x})$.}
    \label{fig:architecture}
\end{figure*}

\section{Local Deep Implicit Functions}
\label{sec:method}

In this paper, we propose a new 3D shape representation, Local Deep Implicit Functions (LDIF).  The LDIF is a function that can be used to classify whether a query point $\mathbf{x}$ is inside or outside a shape.  It is represented by a set of $N$ shape elements, each parameterized by 10 analytic shape variables $\mathbf{\theta_i}$ and $M$ latent shape variables $\mathbf{z_i}$:
\begin{equation}
    \mathrm{LDIF}(\mathbf{x,\Theta,Z}) = \sum_{i\in[N]}g(\mathbf{x},\theta_i) (1 + f(\mathbf{x},\mathbf{z}_i))
\end{equation}
where $g(\mathbf{x},\theta_i)$ is a local analytic implicit function and $f(\mathbf{x},\mathbf{z}_i)$ is a deep implicit function.  Intuitively, $g$ provides a density function that defines a coarse shape and region of influence for each shape element, and $f$ provides the shape details that cannot be represented by $g$. 

Like a typical deep implicit function, our LDIF represents a 3D shape as an isocontour of an implicit function decoded with a deep network conditioned on predicted latent variables.   However, LDIF replaces the (possibly long) single latent code of a typical DIF with the concatenation of $N$ pairs of analytic parameters $\theta_i$ and short latent codes $\mathbf{z}_i$ -- i.e., the global implicit function is decomposed into the sum of $N$ local implicit functions.  This key difference helps it to be more accurate, efficient, consistent, scalable, and generalizable (see Section \ref{sec:evaluation}).\AS{Describe the SIF template.}

\vspace*{1mm}\noindent{\bf Analytic shape function.}
The analytic shape function $g$ defines a coarse density function and region of influence for each shape element.   Any simple analytic implicit function with local support would do.  We use an oriented, anisotropic, 3D Gaussian:
\begin{equation}
    g(\mathbf{x}, \theta_i) = c_i e^{-\frac{\lvert\lvert T_i \mathbf{x}\rvert\rvert^2}{2}} %
\end{equation}
where the parameter vector $\theta_i$ consists of ten variables: one for a scale constant $c_i$, three for a center point $\mathbf{p}_i$, three radii $\mathbf{r}_i$, and three Euler angles $\mathbf{e}_i$ (this is the same parameterization as \cite{genova2019learning}, except with 3 additional DoFs for rotation).  The last 9 variables imply an affine transformation matrix $T_i$ that takes a point $\mathbf{x}$ from object space coordinates to the local isotropic, oriented, centered coordinate frame of the shape element.

\vspace*{1mm}\noindent{\bf Deep shape function.}
The deep implicit function $f$ defines local shape details within a shape element by modulating $g$ (one $f$ function is shared by all shape elements). To compute $f$, we use a network architecture based on Occupancy Networks~\cite{mescheder2019occ}.  As in the original OccNet, ours is organized as a fully-connected network conditioned on the latent code $\mathbf{z}_i$ and trained using conditional batch normalization.  However, one critical difference is that we transform the point $\mathbf{x}$ by $T_i$ before feeding it to the network.
Another critical difference is that $f_i$ only modulates the local implicit function $g_i$, rather than predicting an entire, global function.
As a result, our local decoder has fewer network layers (9 vs. 33), shorter latent codes (32 vs. 256), and many fewer network parameters (8.6K vs 2M) than the original OccNet, and still achieves higher overall accuracy (see Section~\ref{sec:evaluation}).

\vspace*{1mm}\noindent{\bf Symmetry constraints.} For shape collections with man-made objects, we constrain a subset of the shape elements (half) to be symmetric with respect to a selected set of transformations (reflection across a right/left bisecting plane).   These ``symmetric'' shape elements are evaluated twice for every point query, once for $\mathbf{x}$ and once for $S\mathbf{x}$, where $S$ is the symmetry transformation.   In doing so, we effectively increase the number of shape elements without having to compute/store extra parameters for them.  Adding partial symmetry encourages the shape decomposition to match global shape properties common in many shape collections and gives a boost to accuracy (Table~\ref{tbl:ablations}).

\section{Processing Pipeline}

The processing pipeline for computing an LDIF is shown in Figure~\ref{fig:architecture}.  All steps of the pipeline are differentiable and trained end-to-end.   At inference time, the input to the system is a 3D surface or depth image, and the output is a set of shape element parameters $\mathbf{\Theta}$ and latent codes $\mathbf{Z}$ for each of $N$ overlapping local regions, which can be decoded to predict inside/outside for any query location $\mathbf{x}$.  Complete surfaces can be reconstructed for visualization by evaluating $\mathrm{LDIF}(\mathbf{x})$ at points on a regular grid and running Marching Cubes \cite{lorensen1987marchingcubes}.  

The exact configuration of the encoder architecture varies with input data type. We encode a {\bf 3D mesh} by first rendering a stack of 20 depth images at 137 x 137 resolution from a fixed set of equally spaced views surrounding the object.   We then give the depth images to an early-fusion ResNet50~\cite{he2016resnet} to regress the shape element parameters $\mathbf{\Theta}$.   Meanwhile, we generate a set of 10K points with normals covering the whole shape by estimating normals from the depth image(s) and unprojecting randomly selected pixels to points in object space using the known camera parameters.   Then, for each shape element, we select a sampling of 1K points with normals within the region of influence defined by the predicted analytic shape function, and pass them to a PointNet~\cite{qi2016pointnet} to generate the latent code $\mathbf{z_i}$.   Alternatively, we could have encoded 3D input surfaces with CNNs based on mesh, point, or voxel convolutions, but found this processing pipeline to provide a good balance between detail, attention, efficiency, and memory.   In particular, since the local geometry of every shape element is encoded independently with a PointNet, it is difficult for the network to ``memorize'' global shapes and it therefore generalizes better.

We encode a {\bf depth image} with known camera parameters by first converting it into a 3 channel stack of 224 x 224 images representing the XYZ position of every pixel in object coordinates.   We then feed those channels into a ResNet50 to regress the shape element parameters $\mathbf{\Theta}$, and we regress the latent codes $\mathbf{Z}$ for each shape element using the same process as for 3D meshes.

\subsection{Training Losses}

The pipeline is trained with the following loss $L$:
\begin{equation}
    L(\mathbf{\Theta}, \mathbf{Z}) = w_P L_P(\mathbf{\Theta}, \mathbf{Z}) + w_C L_C(\mathbf{\Theta})
\label{eq:loss}
\end{equation}

\paragraph{Point Sample Loss $L_P$.} 
The first loss $L_P$ measures how accurately the $\mathrm{LDIF}(\mathbf{x})$ predicts inside/outside of the ground-truth shape. To compute it, we sample 1024 points near the ground truth surface (set $\mathcal{S}$) and 1024 points uniformly at random in the bounding box of the shape (set $\mathcal{U}$). We combine them with weights $w_i \in \{w_\mathcal{S}, w_\mathcal{U}\}$ to form set $C = \mathcal{U} \cup \mathcal{S}$. The near-surface points are computed using the sampling algorithm of \cite{genova2019learning}. We scale by a hyperparameter $\alpha$, apply a sigmoid to the decoded value $\mathrm{LDIF}(\mathbf{x})$, and then compute an $L_2$ loss to the ground truth indicator function $I(\mathbf{x})$ (see~\cite{genova2019learning} for details):
\begin{equation*}
    L_{P}(\mathbf{\Theta}, \mathbf{Z}) = \frac{1}{|\mathcal{C}|}\sum_{\mathbf{x_i} \in \mathcal{C}} w_i \|\mathtt{sig}(\alpha \mathrm{LDIF}(\mathbf{x_i}, \mathbf{\Theta}, \mathbf{Z})) - I(\mathbf{x_i})\|
\end{equation*}

\paragraph{Shape Element Center Loss $L_C$.}  
The second loss $L_C$ encourages the center of every shape element to reside within the target shape.  To compute it, we estimate a signed distance function on a low-res 32x32x32 grid $\mathbf{G}$ for each training shape.   The following loss is applied based on the grid value $G(\mathbf{p}_i)$ at the center $\mathbf{p}_i$ of each shape element:
\begin{equation*}
    L_{C}(\mathbf{\Theta}) =
    \begin{cases}
        \sum_{\mathbf{\theta_i} \in \mathbf{\Theta}} 
        G(\mathbf{p}_i)^2 & G(\mathbf{p}_i) > \beta\\
        0 & G(\mathbf{p}_i) \leq \beta
    \end{cases}
\end{equation*}

Here, $\beta$ is a threshold chosen to account for the fact that $\mathbf{G}$ is coarse. It is set to half the width of a voxel cell in $\mathbf{G}$. This setting makes it a conservative loss: it says that when $\mathbf{p}_i$ is definitely outside the ground truth shape, $\mathbf{p}_i$ should be moved inside. $L_C$ never penalizes a center that is within the ground truth shape.

It is also possible for the predicted center to lie outside the bounding box of $\mathbf{G}$. In this case, there is no gradient for $L_C$, so we instead apply the inside-bounding-box loss from~\cite{genova2019learning} using the object-space bounds of $\mathbf{G}$.

\section{Experimental Setup}
\label{sec:results}

We execute a series of experiments to evaluate the proposed LDIF shape representation, compare it to alternatives, study the effects of its novel components, and test it in applications.  Except where otherwise noted, we use $N=32$ shape elements and $M=32$ dimensional latent vectors during all experiments. 

\vspace*{1mm}\noindent{\bf Datasets.}  When not otherwise specified, experiments are run on the ShapeNet dataset~\cite{chang2015shapenet}. We use the train and test splits from 3D-R$^2$N$^2$~\cite{choy20163d}. We additionally subdivide the train split to create an 85\%, 5\%, 10\% train, validation, and test split. We pre-process the shapes to make them watertight using the depth fusion pipeline from Occupancy Networks~\cite{mescheder2019occ}.  We train models multi-class (all 13 classes together) and show examples only from the test split.

\vspace*{1mm}\noindent{\bf Metrics.} 
We evaluate shape reconstruction results with mean intersection-over-union (IoU)~\cite{mescheder2019occ}, mean Chamfer distance~\cite{mescheder2019occ}, and mean F-Score~\cite{tatarchenko2019single} at $\tau = 0.01$.  As suggested in~\cite{tatarchenko2019single}, we find that IoU is difficult to interpret for low values, and Chamfer distance is outlier sensitive, and so we focus our discussions mainly on F-Scores.

\vspace*{1mm}\noindent{\bf Baselines.} We compare most of our results to the two most related prior works: Occupancy Networks~\cite{mescheder2019occ} (OccNet), the state-of-the-art in deep implicit functions, and Structured Implicit Functions~\cite{genova2019learning} (SIF), the state-of-the-art in structural decomposition. We also compare to the AtlasNet autoencoder~\cite{groueix2018papier}, which predicts meshes explicitly.

\section{Experimental Evaluation}
\label{sec:evaluation}

In this section, we report results of experiments that compare LDIF and baselines with respect to how well they satisfy desirable properties of a 3D shape representation.

\begin{table}
\setlength\tabcolsep{3pt} %
\vspace{-0.3em}
\hspace{-0.75em}
\resizebox{1.05\columnwidth}{!}{
\begin{tabular}{l|ccc|cccc|cccc} 
\toprule
\multirow{2}{*}{Category} & \multicolumn{3}{c}{IoU $(\uparrow)$}  & \multicolumn{4}{c}{Chamfer $(\downarrow)$} & \multicolumn{4}{c}{F-Score $(\uparrow, \%)$} \\
                          & Occ. & SIF  &   Ours        & Occ.          & SIF    & Atl. &           Ours & Occ. & SIF  & Atl. & Ours \\
\midrule
airplane                  &  77.0  & 66.2 & \textbf{91.2} &         0.16   &  0.44 & 0.17    & \textbf{0.10} &  87.8  & 71.4 & 85.1     & \textbf{96.9} \\
bench                     &  71.3  & 53.3 & \textbf{85.6} &         0.24   &  0.82 & 0.31    & \textbf{0.17} &  87.5  & 58.4 & 76.8     & \textbf{94.8} \\
cabinet                   &  86.2  & 78.3 & \textbf{93.2} &         0.41   &  1.10 & 0.81    & \textbf{0.33} &  86.0  & 59.3 & 71.5     & \textbf{92.0} \\
car                       &  83.9  & 77.2 & \textbf{90.2} &         0.61   &  1.08 & 0.70    & \textbf{0.28} &  77.5  & 56.6 & 74.2     & \textbf{87.2} \\
chair                     &  73.9  & 57.2 & \textbf{87.5} &         0.44   &  1.54 & 1.05    & \textbf{0.34} &  77.2  & 42.4 & 60.7     & \textbf{90.9} \\
display                   &  81.8  & 69.3 & \textbf{94.2} &         0.34   &  0.97 & 0.54    & \textbf{0.28} &  82.1  & 56.3 & 71.4     & \textbf{94.8} \\
lamp                      &  56.5  & 41.7 & \textbf{77.9} & \textbf{1.67}  &  3.42 & 1.57    &         1.80  &  62.7  & 35.0 & 51.1     & \textbf{83.5} \\
rifle                     &  69.5  & 60.4 & \textbf{89.9} &         0.19   &  0.42 & 0.16    & \textbf{0.09} &  86.2  & 70.0 & 85.6     & \textbf{97.3} \\
sofa                      &  87.2  & 76.0 & \textbf{94.1} & \textbf{0.30}  &  0.80 & 0.50    &         0.35  &  85.9  & 55.2 & 70.0     & \textbf{92.8} \\
speaker                   &  82.4  & 74.2 & \textbf{90.3} &         1.01   &  1.99 & 1.31    & \textbf{0.68} &  74.7  & 47.4 & 60.7     & \textbf{84.3} \\
table                     &  75.6  & 57.2 & \textbf{88.2} & \textbf{0.44}  &  1.57 & 1.07    &         0.56  &  84.9  & 55.7 & 67.5     & \textbf{92.4} \\
telephone                 &  90.9  & 83.1 & \textbf{97.6} &         0.13   &  0.39 & 0.16    & \textbf{0.08} &  94.8  & 81.8 & 89.6     & \textbf{98.1} \\
watercraft                &  74.7  & 64.3 & \textbf{90.1} &         0.41   &  0.78 & 0.41    & \textbf{0.20} &  77.3  & 54.2 & 74.4     & \textbf{93.2} \\
\midrule
mean                      &  77.8  & 66.0 & \textbf{90.0} &         0.49   &  1.18 & 0.67    & \textbf{0.40} &  81.9  & 59.0 & 72.2     & \textbf{92.2}  \\
\bottomrule
\end{tabular}
} %
\vspace{0.3em}
\caption{\textbf{Autoencoder results.}  Comparison of 3D-R$^2$N$^2$ test set reconstruction errors for OccNet (``Occ.'')~\cite{mescheder2019occ}, SIF (``SIF'')~\cite{genova2019learning}, AtlasNet (``Atl.'')~\cite{groueix2018papier}, and LDIF (``Ours'') autoencoders.}
\label{tbl:autoencoder}
\end{table}
\begin{figure}
    \centering
    \includegraphics[width=\columnwidth]{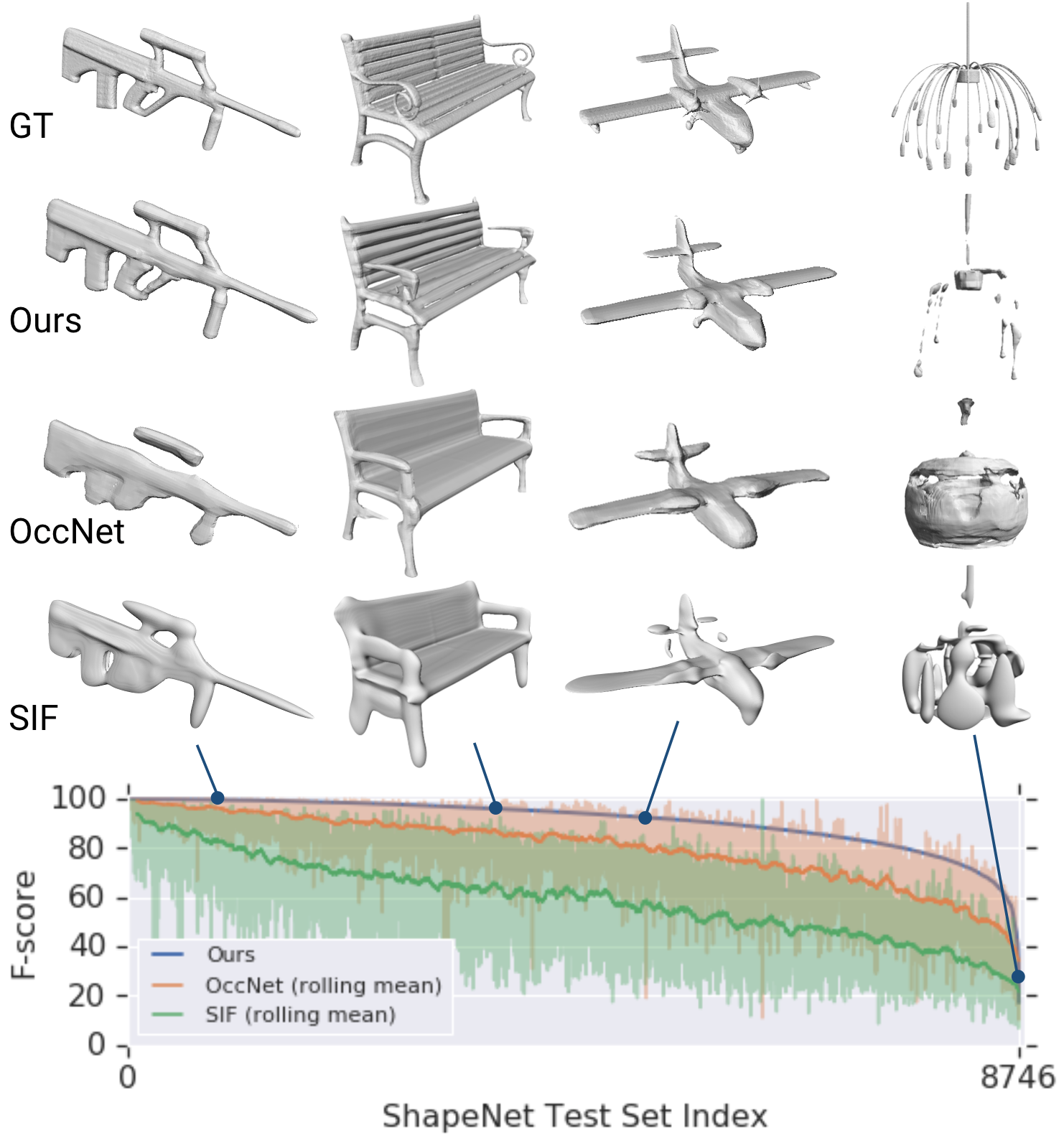} \\
    \caption{\textbf{Autoencoder examples.}  F-scores for the test set (8746 shapes) are shown ordered by the LDIF F-score, with examples marked with their position on the curve. Our reconstructions (blue curve) are most accurate for 93\% of shapes (exact scores shown faded). The scores of OccNet and SIF follow roughly the same curve as LDIF (rolling means shown bold), indicating shapes are similarly difficult for all methods. Solid shapes such as the rifle are relatively easy to represent, while shapes with irregular, thin structures such as the lamp are more difficult.\vspace*{-0.2in}} 
    \label{fig:qual-autoencoder}
\end{figure}

\vspace*{1mm}\noindent{\bf Accuracy.} Our first experiment compares 3D shape representations in terms of how accurately they can encode/decode shapes.  For each representation, we compare a 3D$\rightarrow$3D autoencoder trained on the multiclass training data, use it to reconstruct shapes in the test set, and then evaluate how well the reconstructions match the originals (Table \ref{tbl:autoencoder}). LDIF's mean F-Score is 92.2, 10.3 points higher than OccNet, 20.0 points higher than AtlasNet, and 33.2 points higher than SIF. A more detailed breakdown of the results appears in Figure~\ref{fig:qual-autoencoder}, which shows the F-scores for all models in the test set -- LDIF improves on OccNet's score for 93\% of test shapes.  The increase in accuracy translates into a large qualitative improvement in results (shown above in Figure~\ref{fig:qual-autoencoder}).  For example, LDIF often reproduces better geometric details (e.g., back of the bench) and handles unusual part placements more robustly (e.g., handles on the rifle).

\begin{figure}
\vspace{-1em}
    \centering
    \includegraphics[width=\columnwidth]{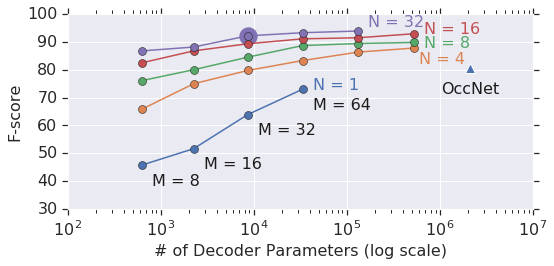}
    \includegraphics[width=\columnwidth]{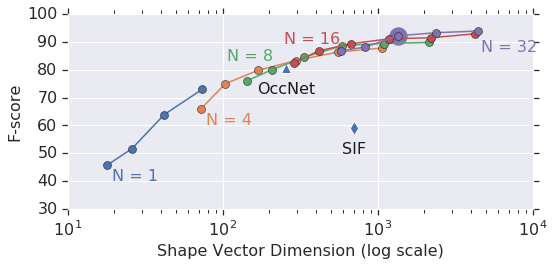}
    \caption{\textbf{Representation efficiency.}  F-score vs. model complexity. Curves show varying $M$ for constant $N$. Other methods marked as points. \textbf{Top}: 
    F-score vs. count of decoder parameters. The $N=32, M=32$ configuration (large dot) reaches \textgreater90\% F-score with \textless1\% of the parameters of OccNet, and is used as the benchmark configuration in this paper. \textbf{Bottom:} F-score vs. shape vector dimension ($\lvert\mathbf{\Theta}\rvert + \lvert\mathbf{Z}\rvert$ for DSIF). DSIF achieves similar reconstruction accuracy to OccNet at the same dimensionality, and can use additional dimensions to further improve accuracy.
    }
    \label{fig:analysis}
\vspace{-1em}
\end{figure}

\vspace*{1mm}\noindent{\bf Efficiency.} 
Our second experiment compares the efficiency of 3D shape representations in terms of accuracy vs. storage/computation costs.  Since LDIF can be trained with different numbers of shape elements ($N$) and latent feature sizes ($M$), a family of LDIF representations is possible, each with a different trade-off between storage/computation and accuracy.  Figure \ref{fig:analysis} investigates these trade-offs for several combinations of $N$ and $M$ and compares the accuracy of their autoencoders to baselines.  Looking at the plot on the top, we see that LDIF provides more accurate reconstructions than baselines at every decoder size -- our decoder with $N=32$ and $M=32$ is $0.004\times$ the size of OccNet and provides $1.13\times$ better F-Score. On the bottom, we see that LDIF performs comparably to OccNet and outperforms SIF at the same number of bytes, despite having both deep and analytic parameters, and that it scales to larger embeddings.

\begin{figure}
    \centering
    \includegraphics[width=\columnwidth]{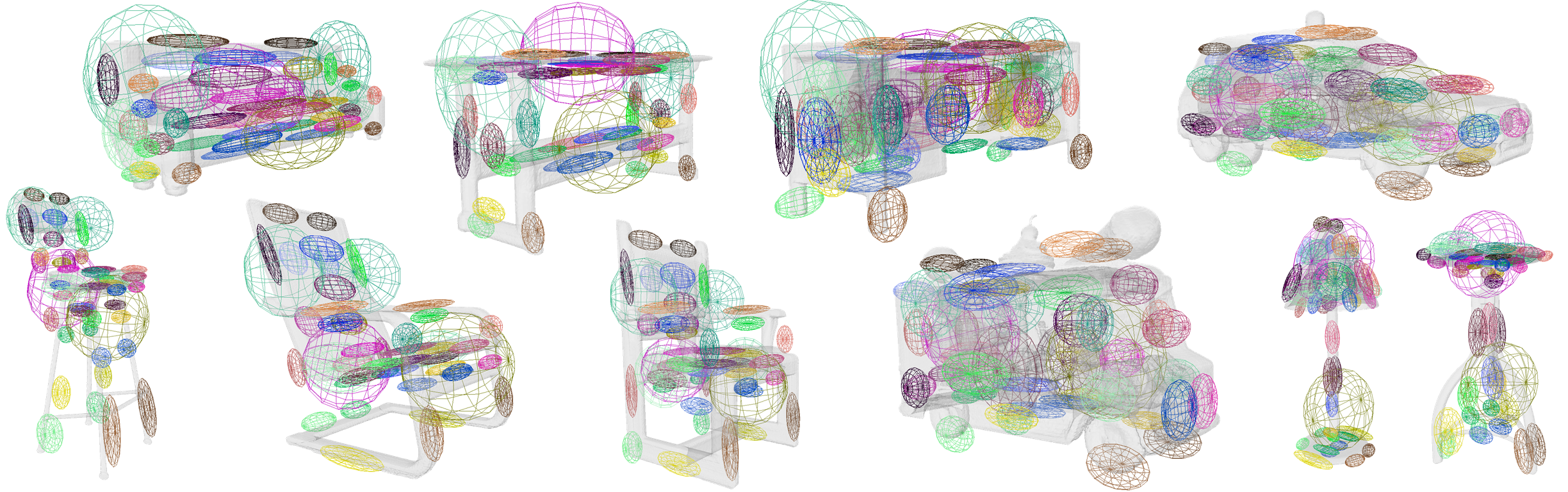}
    \caption{{\bf Representation consistency.}  Example shape decompositions produced by our model trained multi-class on 3D-R$^2$N$^2$.  Shape elements are depicted by their support ellipsoids and  colored consistently by index.   Note that the shape element shown in brown is used to represent the right-front leg of the chairs, tables, desks, and sofas, as well as the front-right wheel of the cars.\vspace*{-0.1in}}
    \label{fig:consistency}
\end{figure}

\vspace*{1mm}\noindent{\bf Consistency.}
Our third experiment investigates the ability of LDIF to decompose shapes consistently into shape elements.   This property was explored at length in  \cite{genova2019learning} and shown to be useful for structure-aware correspondences, interpolations, and segmentations.   While not the focus of this paper, we find qualitatively that the consistency of the LDIF representation is slightly superior to SIF, because the shape element symmetries and rotations introduced in this paper provide the DoFs needed to decompose shapes with fewer elements.   On the other hand, the local DIFs are able to compensate for imperfect decompositions during reconstruction, which puts less pressure on consistency.   Figure~\ref{fig:consistency} shows qualitative results of the decompositions computed for LDIF.  Please note the consistency of the colors (indicating the index of the shape element) across a broad range of shapes.

\begin{figure}
    \centering
    \includegraphics[width=\columnwidth]{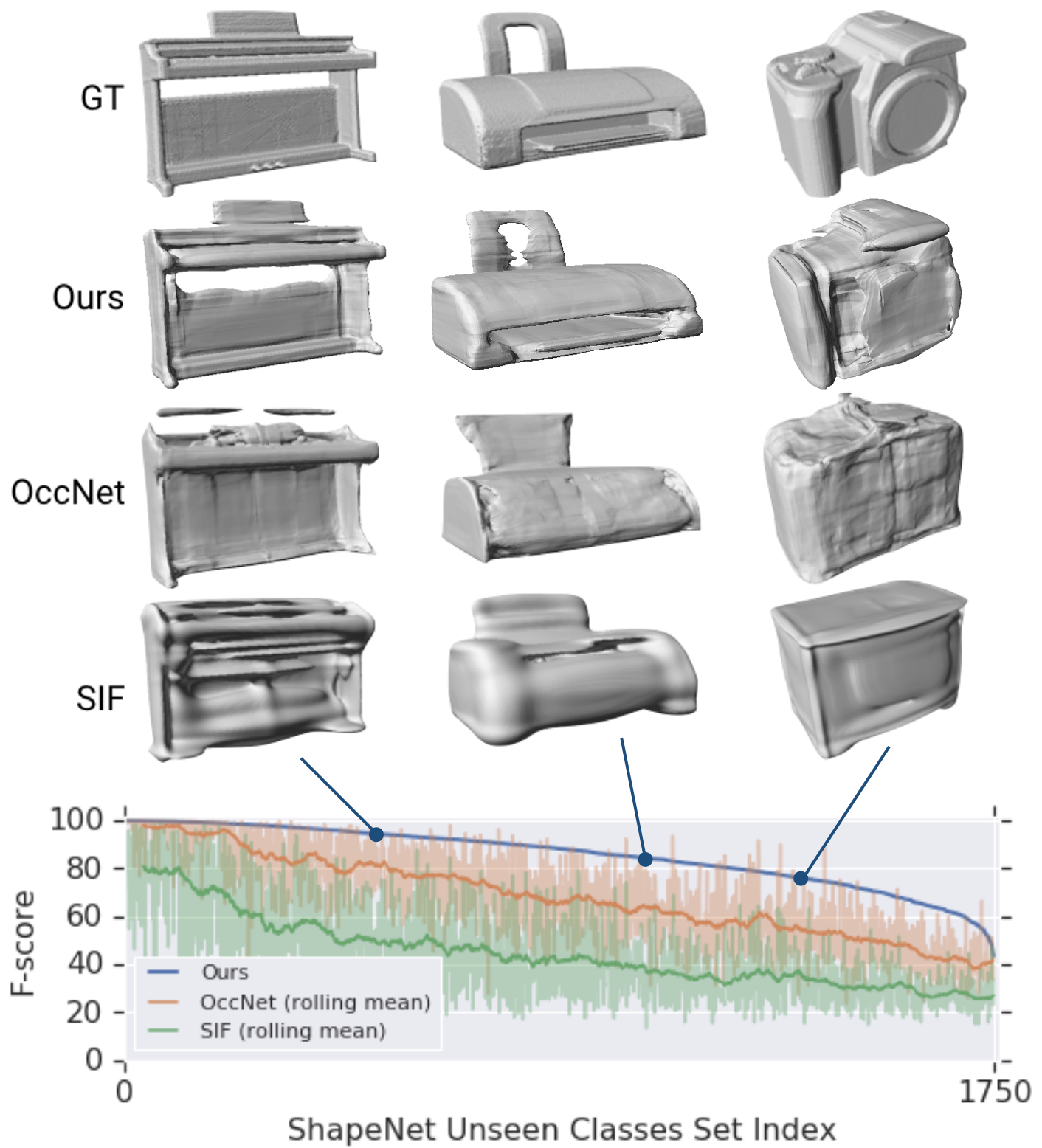} \\
    \caption{\textbf{Generalization examples.} Example shape reconstructions for piano, printer, and camera classes, which did not appear in the training data. F-score is plotted below ordered by LDIF score, similar to Figure~\ref{fig:qual-autoencoder}. Our method (blue curve) achieves the best accuracy on 91\% of the novel shapes.\vspace*{-0.1in}} 
    \label{fig:generalizability}
\end{figure}

\vspace*{1mm}\noindent{\bf Generalizability.}
\label{sec:results-generalizability}
Our third experiment studies how well trained autoencoders generalize to handle unseen shape classes.  To test this, we used the auto-encoders trained on 3D-R$^2$N$^2$ classes and tested them without fine-tuning on a random sampling of meshes from 10 ShapeNet classes that were not seen during training. Table~\ref{tab:generalizability} shows that the mean F-Score for LDIF on these novel classes is 84.4, which is 17.8 points higher than OccNet and 41.4 points higher than SIF.  Looking at the F-Score for every example in the bottom of Figure~\ref{fig:generalizability}, we see that LDIF is better on 91\% of examples.   We conjecture this is because LDIF learns to produce consistent decompositions for a broad range of input shapes when trained multiclass, and because the local encoder network learns to predict shape details only for local regions. This two-level factoring of structure and detail seems to help LDIF generalize.

\begin{table}
\begin{center}
\setlength\tabcolsep{3pt} %
\begin{tabular}{l|ccc|ccc} 
\toprule
\multirow{2}{*}{Category} & \multicolumn{3}{c}{Chamfer $(\downarrow)$} & \multicolumn{3}{c}{F-Score $(\uparrow, \%)$} \\
           & SIF   & OccNet & Ours           & SIF  & OccNet & Ours \\
\midrule
bed        & 2.24 & 1.30  & \textbf{0.68} & 32.0 & 59.3   & \textbf{81.4} \\
birdhouse  & 1.92 & 1.25  & \textbf{0.75} & 33.8 & 54.2   & \textbf{76.2} \\
bookshelf  & 1.21 & 0.83  & \textbf{0.36} & 43.5 & 66.5   & \textbf{86.1} \\
camera     & 1.91 & 1.17  & \textbf{0.83} & 37.4 & 57.3   & \textbf{77.7} \\
file       & 0.71 & 0.41  & \textbf{0.29} & 65.8 & 86.0   & \textbf{93.0} \\
mailbox    & 1.46 & 0.60  & \textbf{0.40} & 38.1 & 67.8   & \textbf{87.6} \\
piano      & 1.81 & 1.07  & \textbf{0.78} & 39.8 & 61.4   & \textbf{82.2} \\
printer    & 1.44 & 0.85  & \textbf{0.43} & 40.1 & 66.2   & \textbf{84.6} \\
stove      & 1.04 & 0.49  & \textbf{0.30} & 52.9 & 77.3   & \textbf{89.2} \\
tower      & 1.05 & 0.50  & \textbf{0.47} & 45.9 & 70.2   & \textbf{85.7} \\
\midrule
mean       & 1.48 & 0.85  & \textbf{0.53} & 43.0 & 66.6   & \textbf{84.4} \\
\bottomrule %
\end{tabular}
\hspace{.1in}
\end{center}
\caption{\textbf{Generalization to unseen classes}.  Comparison of 3D reconstruction accuracy when 3D autoencoders are tested directly on ShapeNet classes not seen during training.  Note that our method (LDIF) has a higher F-Score by 17.8 points.}
\label{tab:generalizability}
\end{table}

\begin{figure}    
    \centering
    \includegraphics[width=\columnwidth]{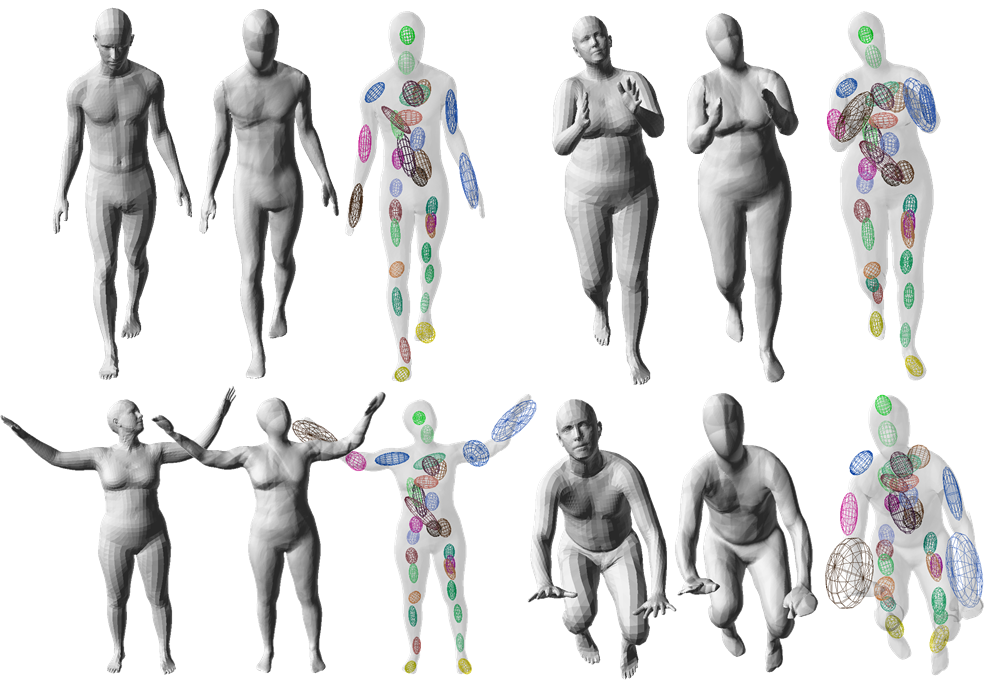}
\caption{\textbf{Human body modeling.}  Surface reconstructions and decompositions for $4$ random SMPL~\cite{SMPL:2015} human meshes from the SURREAL~\cite{varol17_surreal} dataset. For each triple, from left to right: SMPL mesh, our reconstruction, our shape decomposition.  These results demonstrate unsupervised correspondence between people in different poses as well as accurate reconstructions of organic shapes.\vspace*{-0.1in}}
    \label{fig:surreal}
\end{figure}

\vspace*{1mm}\noindent{\bf Domain-independence.}
Our fifth experiment investigates whether LDIF can be used in application domains beyond the man-made shapes found in ShapeNet.  As one example, we trained LDIF without any changes to autoencode meshes of human bodies in a wide variety of poses sampled from~\cite{varol17_surreal}. Specifically, we generated 5M meshes by randomly sampling SMPL  parameters (CAESAR fits for shape, mocap sequence fits for pose). We used an 80\%, 5\%, 15\% split for the train, val, and test sets, similar to~\cite{varol17_surreal}, and measured the error of the learned autoencoder on the held-out test set.  The challenge for this dataset is quite different than for ShapeNet -- the autoencoder must be able to represent large-scale, non-rigid deformations in addition to shape variations.  Our reconstructions achieve $93\%$ mIOU compared to $85\%$ mIOU for SIF\AS{update with MVE, Test}. The results of LDIF reconstructions and the underlying SIF templates are shown in Figure~\ref{fig:surreal}. Despite a lack of supervision on pose or subject alignment, our approach reconstructs a surface close to the original and establishes coarse correspondences. 

\section{Applications}
\label{sec:applications}

In this section, we investigate how the proposed LDIF representation can be used in applications.  Although SIF (and similarly LDIF) has previously been shown useful for 3D shape analysis applications like structure-aware shape interpolation, surface correspondence, and image segmentation \cite{genova2019learning}, we focus our study here on 3D surface reconstruction from partial observations.

\subsection{3D Completion from a Single Depth Image}
\label{sec:applications-depth}

\begin{table}
\begin{center}
\setlength\tabcolsep{3pt} %
\resizebox{\columnwidth}{!}{
\begin{tabular}{l|cc|cc|cc} 
\toprule
\multirow{2}{*}{Category} & \multicolumn{2}{c}{IoU $(\uparrow)$}  & \multicolumn{2}{c}{Chamfer $(\downarrow)$} & \multicolumn{2}{c}{F-Score $(\uparrow, \%)$} \\
                          & OccNet* & Ours & OccNet* & Ours & OccNet* & Ours \\
\midrule
airplane   & - & 80.2 &  0.47 & \textbf{0.17} &  70.1 & \textbf{89.2}  \\
bench      & - & 70.9 &  0.70 & \textbf{0.39} &  64.9 & \textbf{81.9}  \\
cabinet    & - & 82.8 &  1.13 & \textbf{0.77} &  70.1 & \textbf{77.9}  \\
car        & - & 81.4 &  0.99 & \textbf{0.51} &  61.6 & \textbf{72.4}  \\
chair      & - & 70.6 &  2.34 & \textbf{1.02} &  50.2 & \textbf{69.6}  \\
display    & - & 82.4 &  0.95 & \textbf{0.62} &  62.8 & \textbf{80.0}  \\
lamp       & - & 62.1 &  9.91 & \textbf{2.15} &  44.1 & \textbf{66.4}  \\
rifle      & - & 81.5 &  0.49 & \textbf{0.14} &  66.4 & \textbf{92.3}  \\
sofa       & - & 81.4 &  1.08 & \textbf{0.83} &  61.2 & \textbf{71.7}  \\
speaker    & - & 80.2 &  3.50 & \textbf{1.48} &  52.4 & \textbf{67.3}  \\
table      & - & 73.5 &  2.49 & \textbf{1.14} &  66.7 & \textbf{78.0}  \\
telephone  & - & 92.3 &  0.35 & \textbf{0.19} &  86.1 & \textbf{92.0}  \\
watercraft & - & 76.0 &  1.15 & \textbf{0.50} &  54.5 & \textbf{77.5}  \\
\midrule
mean       & - & 78.1 &  1.97 & \textbf{0.76}  &  62.4 & \textbf{78.2}  \\

\bottomrule
\end{tabular}
\hspace{.1in}
}
\end{center}
\caption{\textbf{Depth completion accuracy.}  Our method (LDIF) provides better 3D surface completions than an OccNet* trained on our XYZ image inputs.}
\label{tbl:depth}
\end{table}
\begin{figure}
    \centering
    \includegraphics[width=\columnwidth]{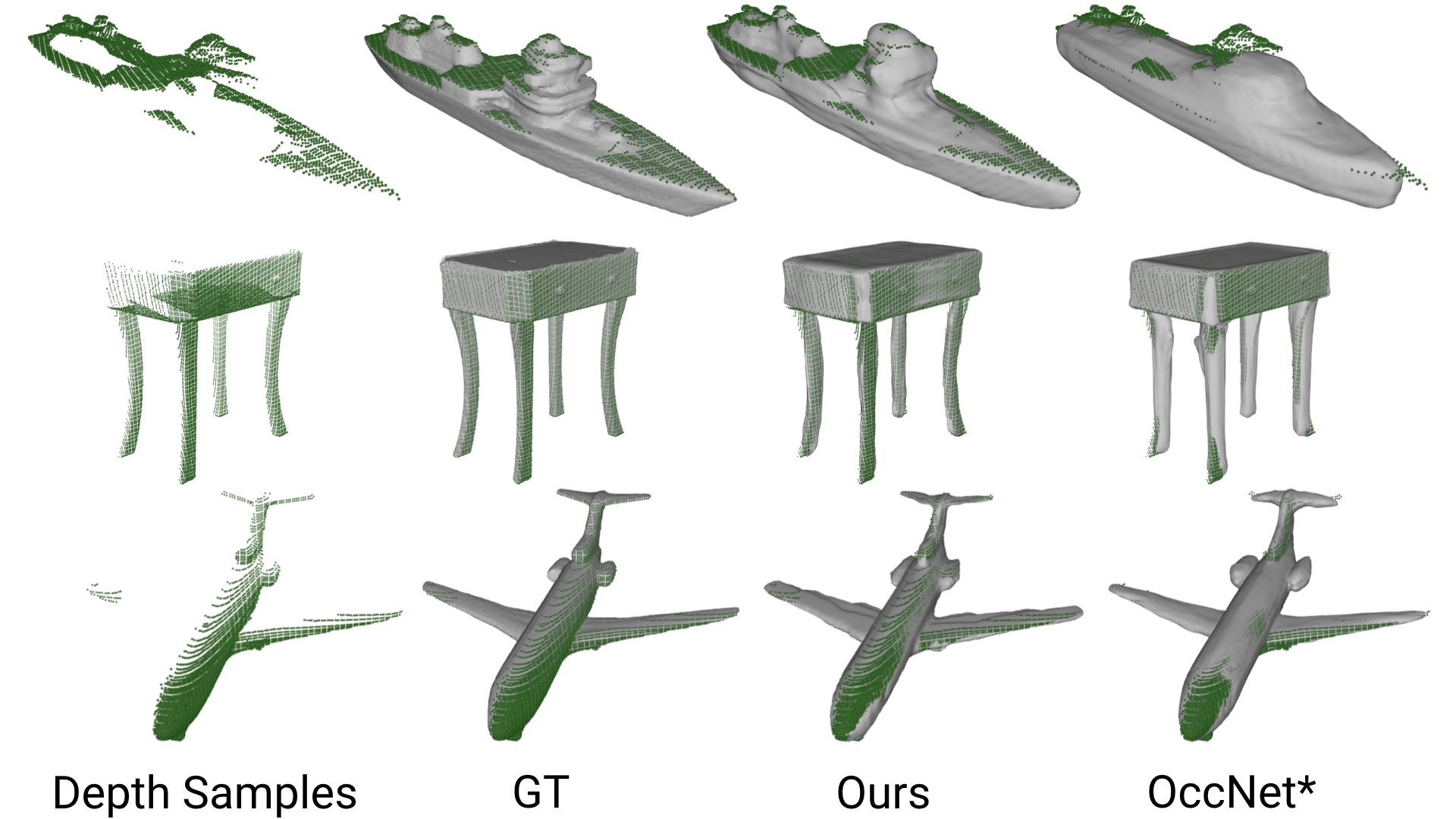}
    \caption{\textbf{Depth completion examples.}  Visualizations of surfaces predicted from posed depth images (depicted by green points). Our method provides better details in both the observed and unobserved parts of the shape.\vspace*{-0.2in}} 
    \label{fig:qual-depth}
\end{figure}

\noindent{\bf Task.}  Reconstructing a complete 3D surface from a single depth image is an important vision task with applications in AR, robotics, etc.   To investigate how LDIF performs on this task, we modified our network to take a single depth image as input (rather than a stack of 20) and trained it from scratch on depth images generated synthetically from random views of the 3D-R$^2$N$^2$ split of shapes. The depth images were 512 x 512 to approximate the resolution of real depth sensors (though all CNN inputs are  224 x 224 due to memory restrictions). The depth images were rendered from view points sampled from all view directions and at variable distances to mimic the variety of scan poses.  Each depth image was then converted to a three channel XYZ image using the known camera parameters.

\vspace*{1mm}\noindent{\bf Baseline.}  For comparison, we trained an OccNet network from scratch on the same data. Because the OccNet takes a point cloud rather than depth images, we train an XYZ image encoder network to regress the 256-D OccNet embedding.  This OccNet* model provides an apples-to-apples baseline that isolates differences due only to the representation decoding part of the pipeline.

\vspace*{1mm}\noindent{\bf Results.}  
Table \ref{tbl:depth} shows results of this 3D depth completion experiment.   We find that the F-Score of LDIF is 15.8 points higher than OccNet* (78.2 vs. 62.4).  Figure~\ref{fig:qual-depth} highlights the difference in the methods qualitatively. As in the 3D case, we observe that LDIF's local part encoders result in substantially better performance on hard examples. 

\vspace*{1mm}\noindent{\bf Ablation study.}  
To further understand the behavior of LDIF during depth completion, we ablate three components of our pipeline (Table~\ref{tbl:ablations}). First, we verify that having local pointnets to encode the local feature vectors is useful, rather than simply predicting them directly from the input image. Second, we show that providing an XYZ image as input to the network is much more robust than providing a depth image. 
Finally, we show that taking advantage of the explicit structure via partial symmetry improves results qualitatively and achieves the same quality with fewer degrees of freedom. The biggest of these differences is due to the PointNet encoding of local shape elements, which reduces the F-Score by 11.4 points if it is disabled.

\begin{table}
\begin{center}
\setlength\tabcolsep{3pt} %
\begin{tabular}{l|ccc} 
\toprule
Method & IoU $(\uparrow)$ & Chamfer $(\downarrow)$ & F-Score $(\uparrow, \%)$\\
\midrule
Full (D) & 77.2 & 0.78 & 77.6\\
No PointNet & 69.1 & 0.98 & 66.2\\
No Transform & 71.9 & 1.80 & 71.9\\
No Symmetry & 76.7 & 0.76 & 76.6 \\
\bottomrule
\end{tabular}
\hspace{.1in}
\end{center}
\caption{\textbf{{\bf Depth completion ablation study.}}
Local PointNet encoders, camera transformations, and partial symmetry all improve performance. Independently and locally encoding the $\mathbf{z}_i$  with PointNet is particularly good for generalization (see Section~\ref{sec:evaluation}). }
\label{tbl:ablations}
\end{table}

\subsection{Reconstruction of Partial Human Body Scans}
\label{sec:applications-caesar}

\begin{figure}
    \centering
    \includegraphics[width=0.95\columnwidth,height=0.55\columnwidth]{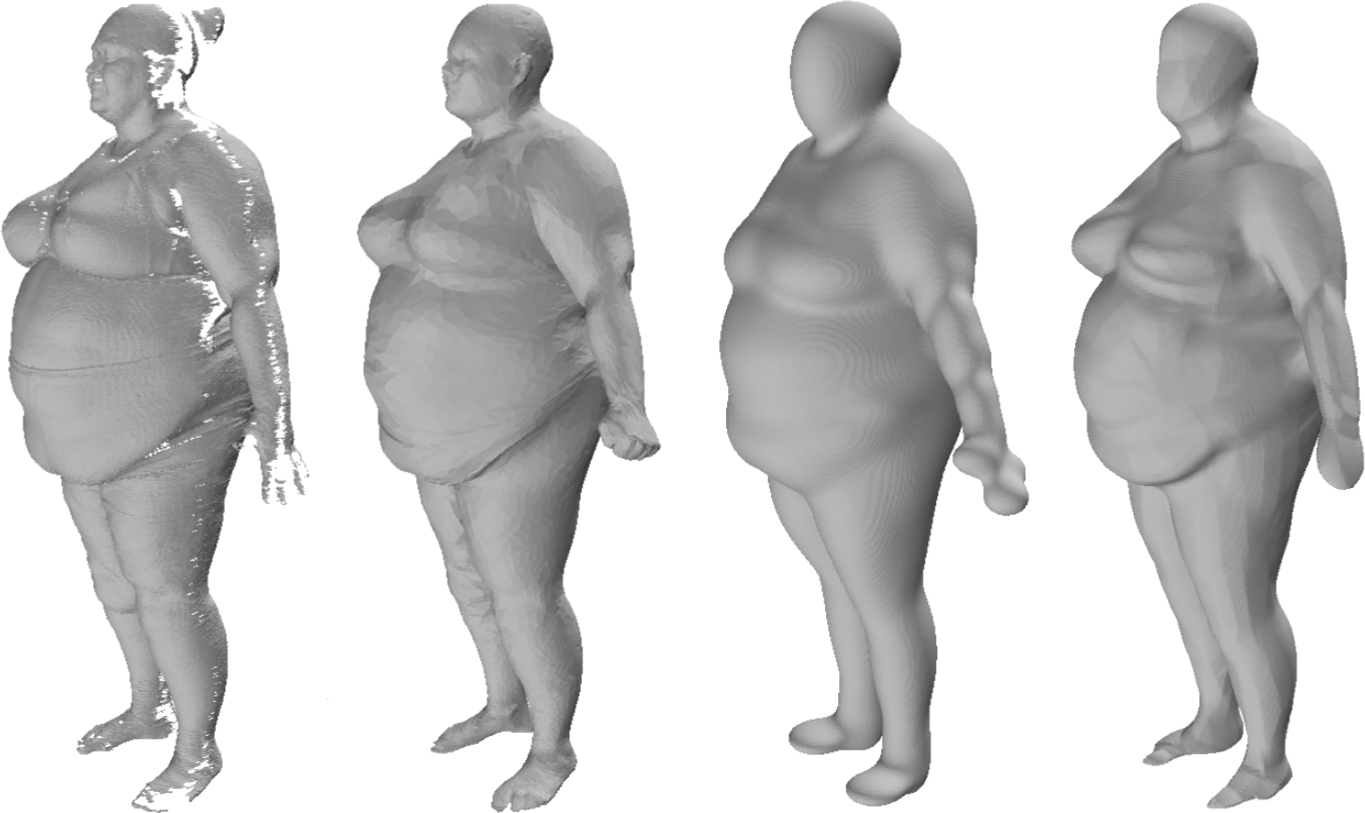}
    \begin{small}
    \hspace{1em}a) Scan~\cite{robinette2002civilian}\hspace{1em}{ b) Template fit~\cite{pishchulin17pr}} \hspace{1em}c) SIF~\cite{genova2019learning} \hspace{1em}  d) Ours\\
    \vspace*{1mm}
    \end{small}    
    \caption{\textbf{Surface reconstruction from partial human scans}.}
    \label{fig:caesar}
\end{figure}

\vspace*{1mm}\noindent{\bf Task.} Acquisition of complete 3D surface scans for a diverse collection of human body shapes has numerous applications \cite{allen2003space}.  Unfortunately, many real world body scans have holes (Figure~\ref{fig:caesar}a), due to noise and occlusions in the scanning process.  We address the task of learning to complete and beautify the partial 3D surfaces without any supervision or even a domain-specific template.

\vspace*{1mm}\noindent{\bf Dataset and baselines.}  The dataset for this experiment is CAESAR~\cite{robinette2002civilian}.   We use our proposed 3D autoencoder to learn to reconstruct an LDIF for every scan in the CAESAR dataset, and then we extract watertight surface from the LDIFs (using the splits from ~\cite{pishchulin17pr}).  For comparisons, we do the same for SIF (another unsupervised method) and a non-rigid deformation fit of the S-SCAPE template~\cite{pishchulin17pr}.

\vspace*{1mm}\noindent{\bf Results.}  Figure~\ref{fig:caesar} shows representative results.   Note that LDIF captures high-frequency details missing in SIF reconstructions.  Although the approach based on S-SCAPE provides better results, it requires a template designed specifically for human bodies as well as manual supervision (landmarks and bootstrapping), whereas LDIF is domain-independent and unsupervised.  These results suggest that LDIF could be used for 3D reconstruction of other scan datasets where templates are not available.

\section{Conclusion}
\label{sec:conclusion}

\noindent{\bf Summary of research contributions:} In this paper, we propose Local Deep Implicit Functions (LDIF), a new 3D representation that describes a shape implicitly as the sum of local 3D functions, each evaluated as the product of a Gaussian and a residual function predicted with a deep network.   
We describe a method for inferring an LDIF from a 3D surface or posed depth image by first predicting a structured decomposition into shape elements, encoding 3D points within each shape element using PointNet \cite{qi2016pointnet}, and decoding them with a small residual decoder.  This approach provides an end-to-end framework for encoding shapes in local regions arranged in a global structure.

We show that this LDIF representation improves both reconstruction accuracy and generalization behavior over previous work -- its F-Score results are better than the state-of-the-art \cite{mescheder2019occ} by 10.3 points for 3D autoencoding of test models from trained classes and by 17.8 points for unseen classes. 
We show that it  dramatically reduces network parameter count -- its local decoder requires approximately 0.4\% of the parameters used by~\cite{mescheder2019occ}.
We show that it can be used to complete posed depth images -- its depth completion results are 15.8 percentage points higher than \cite{mescheder2019occ}.
Finally, we show that it can be used without change to reconstruct complete 3D surfaces of human bodies from partial scans.

\vspace*{1mm}\noindent{\bf Limitations and future work:} Though the results are encouraging, there are limitations that require further investigation.   First, we decompose space into a flat {\em set} of local regions -- it would be better to consider a multiresolution hierarchy.  Second, we leverage known camera poses when reconstructing shapes from depth images -- it would be better to estimate them.  Third, we estimate a constant number of local regions -- it would be better to derive a variable number dynamically during inference (e.g., with an LSTM).  Finally, we just scratch the surface of how structured and implicit representations can be combined -- this is an interesting topic for future research.

\vspace*{1mm}\noindent{\bf Acknowledgements:} 
We thank Boyang Deng for sharing OccNet* training code and Max Jiang for creating single-view depth renders. We also thank Fangyin Wei and JP Lewis for feedback on the manuscript.

\bibliographystyle{style/ieee_fullname}
{\small \bibliography{bibliography.bib}}

\clearpage
\appendix

\section{Hyperparameters}
Table~\ref{tab:hparams} contains all hyperparameter values used for training the model. Architecture details for individual networks are below.

\vspace*{1mm}\noindent{\bf ResNet50.} We use a ResNet50~\cite{he2016resnet} V2 that is trained from scratch. The 20 depth images are concatenated channel-wise prior to encoding.

\vspace*{1mm}\noindent{\bf PointNet.} We modify the original PointNet~\cite{qi2016pointnet} archictecture by removing the 64x64 orthogonal transformation to improve speed and reduce memory requirements.

\vspace*{1mm}\noindent{\bf OccNet.} Our local decoder follows the same overall structure as the original OccNet~\cite{mescheder2019occ}. However, we reduce the number of residual blocks from 5 to 1. The latent layer feature widths are also decreased proportionally to the vector dimensionality.

\vspace*{1mm}\noindent{\bf Local Point Cloud Extraction.} We sample a subset of points for encoding by the local PointNet as follows. We first transform all 10,000 points to the local frame. Then we choose a distance threshold $r=4.0$ measured in local units. Since the local frame is scaled proportionally to the radius, this threshold is approximately four radii in the world frame. We randomly sample 1,000 points without replacement within $r$ and return those as the set of points to be encoded. If 1,000 points do not exist, we expand $r$ until 1,000 total points are found.

\vspace*{1mm}\noindent{\bf Global Point Cloud Creation.} In order to create 10,000 points from one or more input depth images, we randomly sample valid points without replacement from the depth images. If 10,000 valid pixels do not exist, we repeat random points as necessary before moving to the local extraction phase.

\vspace*{1mm}\noindent{\bf Activation Functions.} Since the generated network activations $\mathbf{y}$ are in the range $[-\infty, \infty]$, we apply activation functions to latents $\mathbf{y}$ to interpret them as the analytic parameters $\theta_i$. The following functions are used. For constants $c_i$: $-|y_{c,i}|$. For ellipsoid radii $r_i$: $0.15\times\mathtt{sig}(y_{r,i})$. For ellipsoid euler-angles $e_i$: $\max(\min(\frac{\pi}{4}, y_{e, i}), \frac{-\pi}{4})$. For ellispoid positions $p_i$: $\frac{y_{p,i}}{2}$.

\vspace*{1mm}\noindent{\bf Metrics.} Below we report details for how each metric is computed. All metrics are computed against the watertight version of the ground truth mesh in order to be consistent with the OccNet~\cite{mescheder2019occ} procedure, and are computed in the normalized coordinate frame provided by \cite{mescheder2019occ}. Our predicted results are initially generated in a coordinate frame normalized using the centroid and variance of the mesh, rather than the bounding-box-based normalization of \cite{mescheder2019occ}. Therefore we transform back to the bounding-box normalized frame to compute metrics.

\vspace*{1mm}\noindent{\bf IoU.} 100,000 uniform point samples with inside/outside labels are distributed by \cite{mescheder2019occ}. We evaluate the surface at these locations and compute the IoU between samples that are labeled inside by our representation and samples labeled as inside by the provided points.

\vspace*{1mm}\noindent{\bf F-Score.} F-Score is computed at an absolute threshold of $\tau=0.01$ units in the normalized coordinate space of \cite{mescheder2019occ}. We randomly sample 100,000 points on the surface of each mesh, and use point-to-point distances.

\vspace*{1mm}\noindent{\bf Chamfer.} Chamfer distance is computed with the following formula. The factor of 100 is a scaling factor applied for consistency with existing approaches and readability.

\[
100 * \left(\frac{1}{|A|}\sum_{a \in A} \min_{b \in B} \|a - b\|^2_2 + \frac{1}{|B|}\sum_{b \in B} \min_{a \in A} \|a - b\|^2_2\right)
\]

\makeatletter
\setlength{\@fptop}{5pt}
\makeatother
\begin{table}[t!]
\centering
\begin{tabular}{||c|c||}
    \hline
    Name & Value\\
    \hline
    $\alpha$ & 100.0 \\
    $w_\mathcal{S}$ & 0.1\\
    $w_\mathcal{U}$ & 1.0\\
    $w_C$ & 10.0\\
    $w_P$ & 1.0\\
    Batch Size & 24\\
    Adam $\beta_1$ & 0.9\\
    Adam $\beta_2$ & 0.999\\
    Learning Rate & $5\times10^{-5}$\\
    Surface Isolevel & -0.07\\
    \hline
\end{tabular}
\vspace{1em}
\caption{Hyperparameters and optimization details for training the autoencoder network.}
\label{tab:hparams}
\end{table}

\end{document}